# Using deep learning to enhance electronic service quality: Application to real estate websites

Samaa Elnagar

*Howard University, Information Systems and Supply Chain, Washington, DC, USA*

A B S T R A C T

Electronic service quality (E-SQ) is a strategic metric for successful e-services. Among the service quality dimensions, tangibility is overlooked. However, by incorporating visuals or tangible tools, the intangible nature of e-services can be balanced. Thanks to advancements in Deep Learning for computer vision, tangible visual features can now be leveraged to enhance the browsing and searching experience electronic services. Users usually have specific search criteria to meet, but most services won't offer flexible search filters. This research emphasizes the importance of integrating visual and descriptive features to improve the tangibility and efficiency of e-services. A prime example of an electronic service that can benefit from this is real-estate websites. Searching for real estate properties that match user preferences is usually demanding and lacks visual filters, such as the Damage Level to the property. The research introduces a novel visual descriptive feature, the Damage Level, which utilizes a deep learning network known as Mask-RCNN to estimate damage in real estate images. Additionally, a model is developed to incorporate the Damage Level as a tangible feature in electronic real estate services, with the aim of enhancing the tangible customer experience.

## 1. Introduction

Electronic service quality (E-SQ) is affected by a limited understanding of the service features required for decision-making (Yuan et al., 2013). A principal aspect of electronic services is the visual interface (Ullah & Sepasgozar, 2019). Visual interfaces enhance service representations and the user experience (Kumar et al., 2019). Moreover, visual interfaces enhance the tangibilty of electronic services. *Tangibility* is the visual dimension of the service quality, where the service representation gives a clear, concrete image of the service (Santos, 2002). Videos, demos, and promotional materials are extensively used as tangible tools (Qiqi, 2023). Real estate websites are vital electronic services that represent almost 10 % of the economy (Del Giudice et al., 2017). Though, enhancing the electronic real-estate service quality received little attention (Jarosz et al., 2020).

According to Zillow (Chen et al., 2011), almost half of real estate consumers regret their decision to purchase or rent a property. The primary reasons for these regrets are the limited information about the properties (Ullah et al., 2018). Furthermore, a study on the quality of real estate services found that images are the most crucial aspect of real estate websites (Choi & Kim, 2017; Ullah & Sepasgozar, 2019). These images capture important structural and visual features, such as the

Damage Level. The Damage Level is a semantic feature that represents to what extent are the building sections damaged? The Damage level is rated from 1 (extreme or severe damage) to 4 (no damage), on a rating scale (Ciuna et al., 2017). Extracting information from images is considered one of the ``Tangibles'' that significantly enhances customer satisfaction (Panda & Das, 2014).

This research uses deep learning for computer vision to detect the damage level of properties on real estate web sites. The *Damage Level* is a visual feature and a search filter that is considered a tangible tool in one of the most vital e-services which is real estate websites. The research assigns each property a "*Damage Level*", which is used to filter properties based on the state of maintenance. The *Damage Level* is extracted automatically from real estate images using the deep learning network of the Mask Region-based Convolutional Neural Network (Mask-RCNN) (He et al., 2017; Nie et al., 2020).

The research has many social impacts, and it emphasizes the effect of tangible tools on enhancing information representation and lowering the cognitive load in decision making using electronic services. The research also contributes to the vital economic sector of real estate by introducing the *Damage Level* as visual feature and research tool. Tangible tools are desperately needed in many other sectors, such as logistics, tourism, and beauty (Chen, 2023).

E-mail address: smaazoon@gmail.com.

The subsequent sections provide an overview of previous studies on e-service quality. Subsequently, we discuss the detection of the *Damage Level* feature. Following that, the methodology of the proposed system is presented. The proposed model is described and evaluated through demonstrative examples. Lastly, prospect work is addressed.

## 2. Analysis of literature

*E-services that are* selling goods and services lack physical engagement. Therefore, *Tangibles* were considered to be the most important dimension in increasing customer loyalty and trust (Gefen, 2002). *Tangibility* is most closely linked to the physical appearance of facilities. For electronic services, most research focused on the visual appearance of the websites (offline scale) but little research addressed using other visual tools (dynamic scales), such as graphics and highlights of important information in images to enhance tangibles (Parasuraman et al., 2005). Several studies have also realized the strong link between the application of artificial intelligence methods and the improvement in comprehending real estate websites, as well as enhancing user satisfaction and service quality (Winson-Geideman & Krause, 2016). Ullah and Sepasgozar (2019) validated the absence of the dynamics that regulate the integration of information technologies in the real estate industry, which had an impact on user experience.

### 2.1. Earlier scholarly work

Liu et al. (2019) used deep learning for computer vision to assess the aesthetics and quality of portal images in e-retailers, which depend on images of products and services. The study found that enhancing image quality increases willingness to purchase and customer satisfaction. Qiqi (2023) study examined the effect of tangibility and other SQ measures on purchasing willingness, and confirmed the importance of tangible tools to provide clear and detailed information about products or services.

Several online APIs in the real estate industry, such as restb.ai.,[1] can assess real estate properties. However, these APIs lack criteria that explain the evaluation process and the allocation of damage to different components of the property. This absence of clear criteria leads to highly inaccurate evaluations. To address this issue, it is crucial to localize and quantify each image, which will help users understand the reasoning behind assigning specific conditions to the entire property.

There has been little research on estimating the visual damage conditions of properties, even though determining the damage condition is a significant concern in real estate. A deep learning system was created by a study (Poursaeed et al., 2018) using images level-of-prestige. The system incorporates the degree of leisure and additional factors into the value evaluation process. To determine the degree of leisure in each area of the property, the authors used DenseNet networks. To assess the true condition of a house, however, relying solely on the leisure level is inadequate. In a different study, Elnagar and Thomas (2019) combined Automated Valuation Models (AVM) for estimating property prices with Mask R-CNN to estimate property damage conditions. Their system primarily aimed to enhance AVM performance for real estate agents, without providing customers with the ability to filter properties based on damage conditions. To summarize, tangible tools have a significant impact on improving the quality of electronic services and customer satisfaction (Ullah & Sepasgozar, 2019). This emphasizes the need for more comprehensive tangible filters within real estate e-services to enhance information representation and the relevance of search results, enabling informed decision-making in real estate (Limsombunchai, 2004).

## 3. Theoretical framework

E-SQ has psychometrical dimensions where some features are more effective to some users than others (Theodosiou et al., 2019). In electronic services that include physical products and services, the E-SQ dimensions of *information quality* and *Tangibility* are highly correlated with user satisfaction (Dabholkar & Overby, 2005; DeLone & McLean, 2003; Zhao et al., 2012). According to the *Visual Metaphor Theory* (Refaie, 2003), visual metaphors enhance the comprehension and retention of abstract concepts and turn into them more concrete and tangible concepts. Real estate images are rich with visual metaphors about property condition, yet this information is hard to use in searching and filtering. In addition, the *Cognitive Load Theory* (Plass et al., 2010) confirms the effect of visual representations that are more tangible in reducing the cognitive load needed to make decisions. Another Neuro-IS study found that images that provide expressive cues are related to the level of information recall, especially when images are *relevant* to the information provided (Riaz et al., 2018).

Real estate is a unique one-off purchase product that is usually immovable. Real estate e-services lack quality, detailed property photos that might negatively impact user's perception about the real estate websites' (Ullah et al., 2018; Dabholkar et al., 2000; Ali et al., 2021). Liu et al. (2019) asserts the importance of visually oriented communication, especially pictorial components, to build association, enhance decision-making, and overcome the intangible and perishable nature of real estate e-services.

The introduction of new AI technologies enhances the tangibles that businesses use in service delivery models (Omoge et al., 2022). If real estate images could indicate the damages to a property, real estate websites could filter them and allow users to select the *Damage Level* accepted. Moreover, locating damages in real estate photos is considered a visual tool that would enhance the tangibility of a real estate web site. In this research, deep learning is used to identify and quantify damage in real estate images as a tangible tool. The main service quality issues that are addressed in the research context and details about each measure are provided below.

*Efficiency:* refers to the reduction of search costs, the availability of information, and the ability to make multi-attribute assessments (Xue et al., 2000). *Efficiency* directly affects the outcome of the service and the quality of the information (Zhao et al., 2012). In this study, we measure efficiency by comparing the number of listings a user must browse using the proposed model versus conventional real estate websites in order to find a desired listing.

*Tangibility:* measure how tangible resources affect service to customers (Pakurár et al., 2019; Sun et al., 2012). A significant quality issue brought up by Sun et al. (2012) is addressed by the *Damage Level*. Which answers the question of whether the pictures of the property offer satisfactory details to the website user? For the purpose of assisting in real estate decisions, the Damage Level aims at localizing and quantifying damage in each property image. Damage level improves information representation and provides a tangible user experience (Seiler & Reisenwitz, 2010).

## 4. Methodology

This research is following a design science approach based on Elnagar and Thomas (2019) research, in which the Mask R-CNN deep learning network was used for estate appraisal based on property images. The network is trained to identify different *Damage Level* as described below (Offermann et al., 2010). The trained Mask RCNN is considered an instantiation artifact that is evaluated using illustrative examples.

### 4.1. Damage level as a tangible feature

The *Damage Level* is a weighted composite attribute inspired by the

---

[1] RestAI

*State of maintenance (STM)* measure. STM was introduced by Ciuna et al. (2017) and represents the level of physical degradation of a building's internal and external components. As indicated in Fig. 1, each section has a significance weight that gets evaluated by real estate professionals based on its susceptibility to damage. Upper-floor bathrooms, for example, are more prone to leaks than lower-level restrooms. As a result, their significance weights are higher. Each section or room is made up of components such as floors and walls, and each component is assigned an importance weight based on its cost of maintenance. Because of the higher expense of maintenance, ceilings are normally given that most importance weight.

On the ordinal scale, there are four levels of damage: 1 severe damage or complete damage, 2 mild damage or moderate level of wear and tear, 3 minor damage or marginal normal wear, 4 None or a satisfactory state of repair. However, if a component, for example: the floor, has many damages of varying severity. The component level of damage is the highest damage Level detected. For instance, suppose the floors have two mild defects and one severe damage. A severe level is assigned to the floors. A section/room damage Level is calculated as in Eq. (2) as the average weighted damage of the section's components:

$$CD = \max(d)_{1 \leq i \leq 4} \tag{1}$$

$$Sec_i = \frac{1}{n} \sum_{j=1}^{n} \left( W_j \ CD_j \right) \tag{2}$$

where $d$ is the *Damage Level* detected in certain component. *CD* is the component's Damage Level. *Sec_i* is the room/section assigned Damage Level based on the average damage of its components $n$, and each component $j$ has a weight of $W_j$ based on its importance.

## 5. The proposed model

The proposed model suggests how to incorporate the Damage Level in real estate websites as Fig. 2 illustrates. The model is comprised of three major modules: damage detection, Damage Level assignment, and a search and filtering process. The next section goes into detail about each module:

### 5.1. Damage allocation detection

The primary role of this module is to identify the damages present in the images of each property. Once a realtor uploads new images of a property into the system, the module automatically runs damage detection on each image. In addition, if a user submits a search or filtering request and some of the retrieved properties have not undergone damage detection, those properties will be prioritized for damage detection.

#### 5.1.1. Image correction

To ensure consistency, each image is corrected as a basic preprocessing step. Image correction includes removing blur, adjusting contrast, and adjusting brightness. In addition to rotation adjustments, and skew correction, However, images with low resolutions or pixel dimensions of less than 72 PPI are rejected and excluded from processing.

#### 5.1.2. How Mask-RCNN identify damage

Mask-RCNN network is used to detect damage in real estate images. It has two main components, the backbone network and a head network. The backbone network is a faster R-CNN (ResNet) which uses the Feature Pyramid Network (FPN) to build a bottom-up pyramid-like feature map. FPN are known for maintaining strong semantic features at various resolution scales. FPN therefore is excellent choice to learn different degrees of damage features such as paint cracks, water damage and physical damages (He et al., 2017).The feature map is then fed to

Region Proposal network (RPN) to generate Region of Interest (ROI). However, ROI is bonded to its raw image. So, a special module called ROI Align and turn feature maps into a fix-sized tensors (Lin et al., 2017) as in Fig. 3.

The head network take ROI align output and passes them through a set of fully convolutional network (FCN) to generate the damage masks. The same ROI outputs are passed through fully connected networks (FC) to generate each damage class and the bounding box surrounding that damage. Additionally, the network provides a confidence score for each detected damage, as shown in Fig. 4. The confidence score reflects the network's certainty in assigning a particular damage level to an identified damage (Lu et al., 2020).

### 5.2. Damage level estimation

This module is responsible for estimating the property Damage Level based on the damage detected in the uploaded photos and saving the estimated Damage Level to the property database.

#### 5.2.1. Property data model

To better understand how the Damage Level will be used in the real estate data records, the basic data model entities are detailed below. The full data model is in this appendix.[2]

1. **Property**: it stores the basics of the property, like zip code and build year and *total_level_damage*.
2. **Property_Criteria**: it shows the values of each of the properties' criteria. Sec_*Sign_Weight*: it shows how vulnerable a section is to damage, and *sec_Level_Damage* which is the total of the damage from the section's components.
3. **P_Components**: every section or room has a number of components such as the *windows, floors and walls*. Each component has a comparative weight *comp_imp_weight* that is based on its cost of maintenance. When the mask RCNN identify a damage, it is saved in the *comp_level_damge* attribute.
4. **P_Sections**: represents the sections of a property, such as kitchen, are assigned a significance score (*sec_sign_weight*) that indicates the section's vulnerability to deterioration. The *sec_level_damage* value is the section's Damage Level, which is derived from the total of the damage to its components.

The *Damage Level assignment* is firstly performed on the component level then on the section level as described Eqs. (1) and (2). Based on the Damage Level of all sections, the entire property is assigned a *Damage Level*. So, the property damage *PD* is simply the average damages of all property sections as represented in the following equation.

$$PD = \sum_{i=1}^{m} W_{li} \ Sec_i \Big/ (m) \tag{3}$$

where $i$ rangs from 1 to $m$ representing the number of sections in a property, and each section $Sec_i$ has a significance weight $W_{li}$. This $W_{li}$ is determined by real estate professionals depending on each section's significance. For example: a garage has less significance weight than a bathroom. Algorithm A summarizes the damage estimation process.

### 5.3. The prototype website

This module serves as the system's interface with users of the real estate website, where it receives their search and filtering requests. The users could filter properties according to their tolerable damage level. The interface also allows displayed properties to be sorted based on their *Damage Level* in ascending or descending order.

---

[2] Data Model Appendix

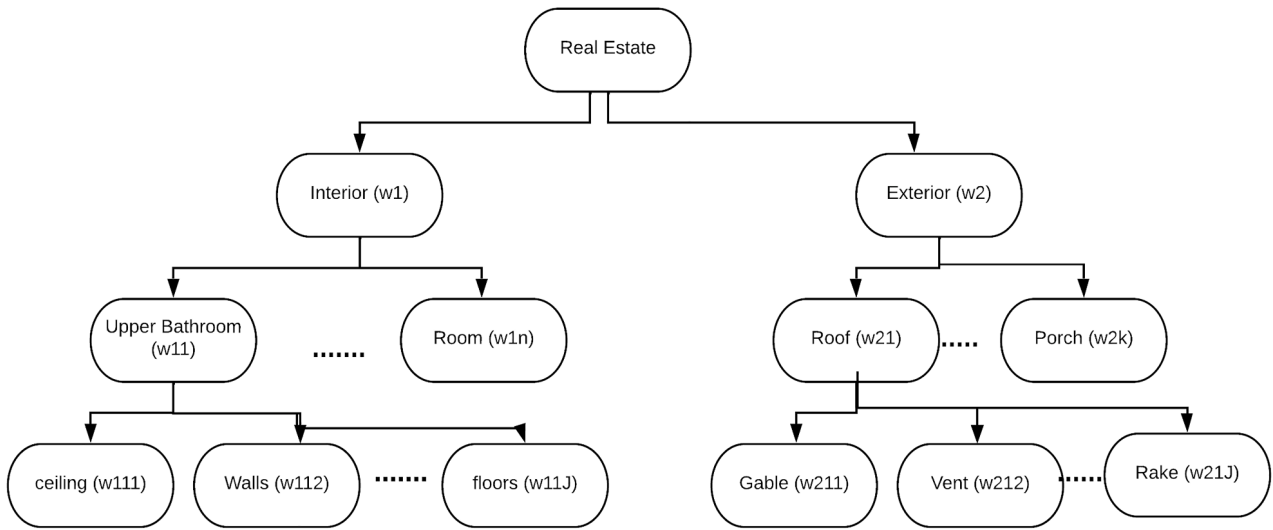

**Fig. 1.** Damage level in different sections and components.

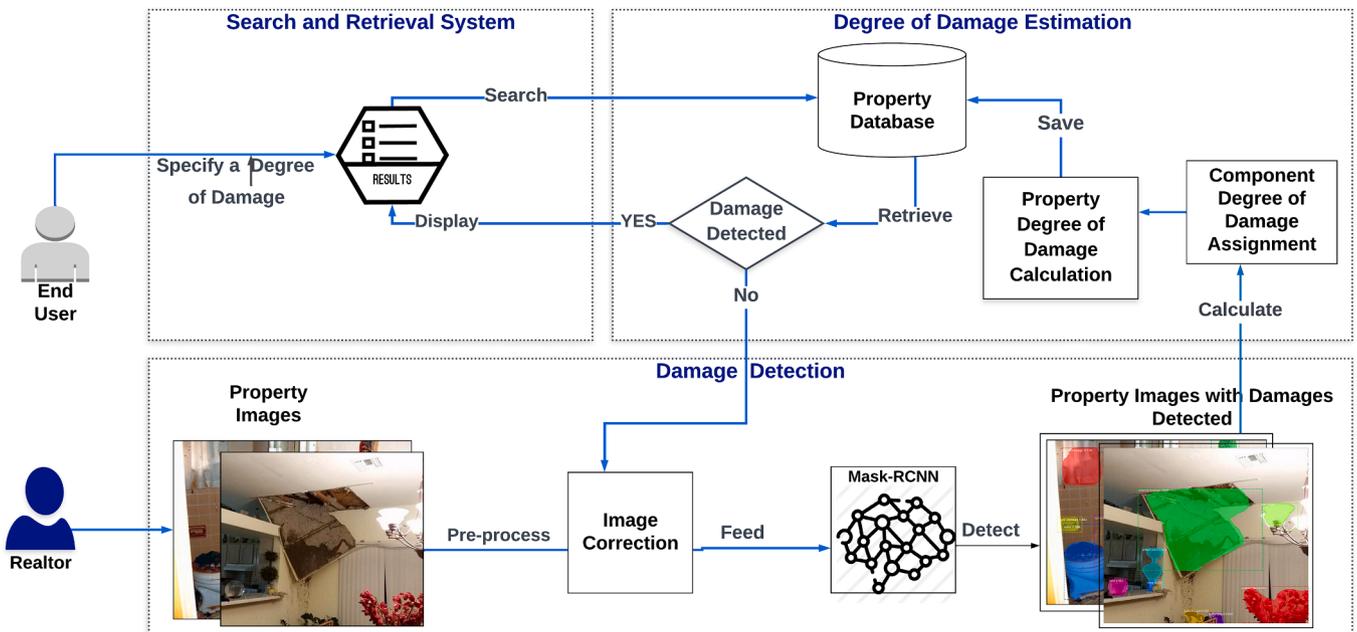

**Fig. 2.** The proposed model architecture.

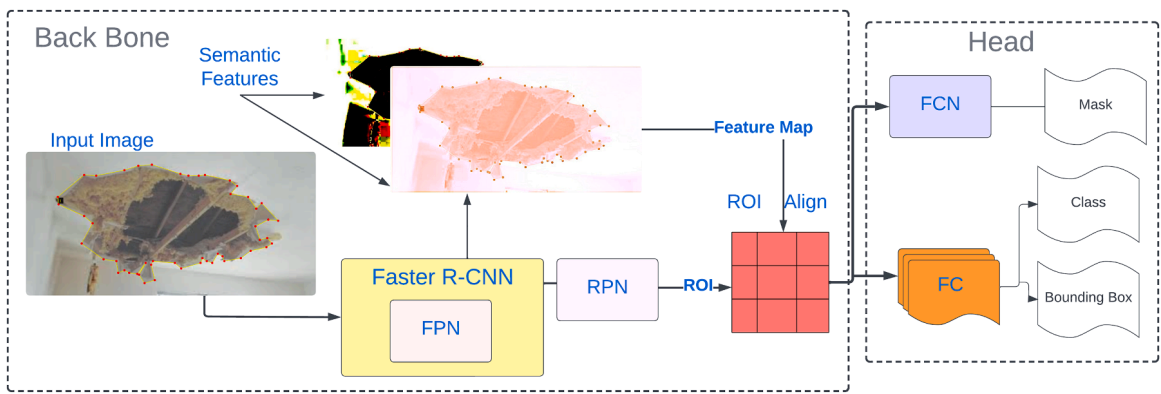

**Fig. 3.** The Mask RCNN network architecture used in detecting damages.

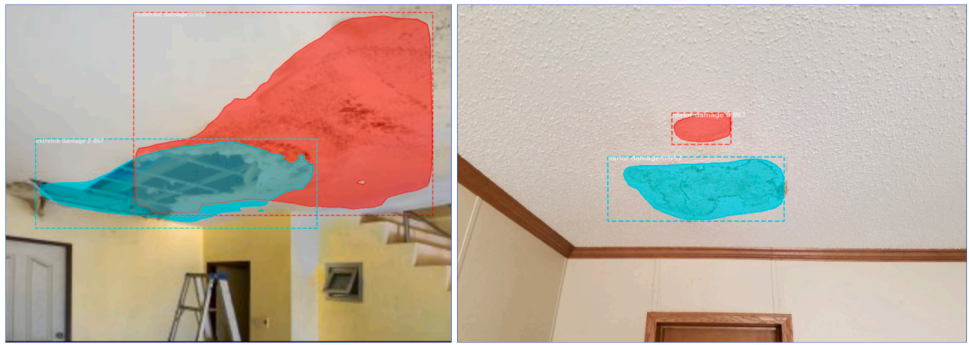

**Fig. 4.** Damage localization along with the surrounding bounding box and confidence score.

**Algorithm A**

Damage level estimation.

| Stage | Explanation |
| --- | --- |
| 1- Examine if damage estimation is needed | Determine if there is a property damage level estimate. If there is, proceed to Step 4; if not, proceed to Step 2 and Step 3. |
| 2- Schedule for Estimation | Images of undetected property are assigned for damage detection. |
| 3- Detect each *image* | Each image consists of one or more components. Also, multiple images can represent the same section. The Mask-RCNN model mask will identify to which section each image belongs, then detect damage in the $CD_j$ component of each image to save as *comp_level_damage*. |
| 4- Estimate damage level of section *section_level_damage* | Compute the section damage level *section_level_damage* based on the *comp_level_damage* ($CD_j$) & the importance weight *comp_imp_weight* ($W_j$) of the section *components* as in Eq. (2). |
| 5- Assign Damage Level to the entire property *PD* | For each property section, replicate process 4. Then, as in Eq. (3), determine the total damage *PD* based on the weighted average *sec_sig_weight* ($W_{li}$) and Damage Level of each property section *sec_level_damage* ($Sec_{li}$). *PD* is saved as a *total_level_damage* in the *Property* entity. |

## 6. Proposed model implementation

The proposed model's implementation is broken down into two sections, the first of which involves training a Mask-RCNN damage level detection network and testing it using photos with different levels of damage. The implementation of the search/filtering website, which applies damage level and sorts the search/retrieval results in accordance with user preferences, is the second stage.

### 6.1. Mask-RCNN damage level training

This step involves training A Mask-RCNN to identify damage in real estate photos. Below is a detailed explanation of the training and testing procedure, including the data selection, network construction, defining criteria for identifying each damage level, annotating training and validation photos, and results.

### 6.1.1. Data selection and image preprocessing

Deep learning networks typically require a large number of training and validation samples. However, there is currently no existing dataset crafted specifically for detecting damage in real estate properties. As a result, we manually downloaded training and testing images from search engines and relevant websites. Given our limited computational resources, we focused on training the network to detect damage, specifically in the *ceiling component*. We obtained images showing various levels of ceiling damage, totaling 594 in number. We carefully selected high-quality images that clearly displayed observable damages, while avoiding images that included unrelated objects like speakers, alarms, and routers. To prepare the images for analysis, we resized them to dimensions of $712 \times 712$ and applied contrast stretching to enhance image contrast (Munteanu & Lazarescu, 1999).

### 6.1.2. Damage level criteria

Quantifying the *Damage Level* for each component of a property can be a challenging task. Often, there are indefinite images that can correspond to multiple levels of damage. To address this, we have developed criteria that allow us to distinguish between different levels of damage in ceilings. By employing these distinct categories, we can accurately assess the varying levels of damage in different types of ceilings.

**Minor class:** This category includes small water stains that range in color from beige to light brown, as well as small paint cracks that result in an uneven surface. Additionally, small moldy spots, usually dark in color, are considered part of this class.

**Mild class:** This class encompasses larger water stains, medium-sized paint cracks, and moldy spots that are larger and more prominent than those in the minor damage class.

**Severe class:** Here, we identify significant damage to the ceiling boards, such as visible signs of deterioration or crumbling sections. This class also includes large areas affected by mold and instances where parts of the ceiling have fallen off. 4.

**No damage class:** This class represents standard ceilings that exhibit no damage, mold, or stains. However, it is important to note that during training, we utilized images of plain ceiling structures to prevent the detection of false positives that could arise from intricate designs or patterns.

### 6.1.3. Annotation

In order to annotate training photos with different levels of damage, we employed VGG annotation tool.[3] Each damage is encircled by a polygon that was depicted as a collection of (x,y)-valued points. Then a class was given to each polygon. Each polygon is represented as a class in a JSON file export as shown.

[`Damage1.jpg`:[`filename`:`Damage1.jpg`,`size`:6122,`regions`:[`shape_attributes`:[`name`:`polygon`,`all_points_x`:[123,128,242,184,143],`all_points_y`:[104,12,70,96,102]],`region_attributes`:[`name`:`no damage`]]]]

### 6.1.4. Network training and testing

The selection of the training and test images is a crucial stage in training any deep learning network. 30 % of the dataset was utilized for validation, 20 % for testing, and 70 % for training. We intended to divide the downloaded photographs into the following categories for training and testing: 30 % Severe, 30 % Mild, 30 % Minor, and 10 % No damage. Due to the difficulties in locating appropriate photographs, the division percentages were slightly off.

---

[3] VGG Annotation

### 6.1.5. Network configuration

The experiments were carried out using Google Collaboratory Pro. Python was utilized to develop the training and testing code. During both training and testing, a single GPU was utilized, with two images processed at a time. The backbone network used a residual learning framework (resnet101), and the default configurations of the Mask-RCNN were followed. A learning rate of 0.1 % and a learning momentum of 9 % were both chosen. Every iteration of the batch size included two images. Images were cropped to fit the $[2^{10} \ 2^{10} \ 2^{10} \ ]$ (Kumar et al., 2019) dimensions to ensure consistency in the results. Images with four channels were downsized to just three. Each training epoch had 100 steps in it. Each step took 119.87 s on average. The average training loss was calculated to be 0.7915, and the average validation loss was 0.7043.

### 6.1.6. Training phases

To streamline the training procedure, we transferred the learning of the weights from the previous pre-trained R-CNN Mask model trained on a dataset known as the MS- COCO[4] dataset. MS-CCOO is a widely used dataset for object detection and pattern detection, as well as for captioning tasks. The dataset contains over 1.5 M object instances. Nevertheless, we modified the training process by substituting the final layer in the pre-training network with the MaxPooling and projection layers. MaxPooling reduces the output dimension, while projection enhances the object annotation capabilities. The training process was broken down into three successive steps, which we will discuss in greater detail below.

### 6.1.7. Pilot test to detect severe damage

As a pilot experiment, we trained the Mask-RCNN network to identify a specific type of damage in the ceiling component, namely severe damage. This experiment aimed to verify that the network could accurately identify the damaged area using instance segmentation and its surrounding bounding box. Fig. 5 illustrates this successful allocation of the damaged mask. During the training process, we re-trained the pre-existing Mask-RCNN model from the COCO dataset. In total, 92 images were employed for the pilot test training, while 18 were reserved for validation, and six images were used for testing. Throughout the process, we adhered to the default confidence level of 0.9, as specified in the original Mask-RCNN code. The network demonstrated successful detection of severe damages and accurately located the corresponding bounding box and binary mask.

### 6.1.8. Complete training of the Mask-RCNN

Following the successful execution of the pilot test, we proceeded to impart training to the network for the identification of four distinct categories of ceiling damage: Severe, Mild, Minor, and None. Where an image might contain more than one damage level, a total of 594 levels of damage instances were used in training. Additionally, we evaluated the network's ability to accurately distinguish between undamaged and damaged ceilings by subjecting it to images devoid of any structural damage.

### 6.1.9. Including other objects while training the Mask-RCNN

In real estate photos, furniture and appliances help to identify the section of the image, so it is important that the network can differentiate between damages and objects. For example, an image that shows a stove represents the kitchen section, so we decided to not train the same network for appliances, light fixtures, kitchen cabinets, etc. we opted for a different approach to avoid biased training due to an imbalance in training samples. So, We trained a separate network for measuring the damage level, while using another network for detecting other objects in the real estate image. As a result, we dedicated a separate general object pre-trained Mask-RCNN with the COCO dataset as a versatile network. Samples of damage detection along with other objects are shown in Fig. 6.

### 6.2. Assigning damage level

The Mask RCNN detection output is a list of identified classes (including damages) along with their confidence scores. The output also includes the identified objects' masks and their bounding boxes. The class IDs of the identified damages range from 1 for the severe damage to 4 for the no-damage. To set the maximum damage identified in the image as discussed in Eq. (1), we simply select the lowest detected class ID. For example, in the two pictures on Fig. 6, there are more than 10 objects identified with 10 class IDs. The lowest class ID in both images is 1 as both images have severe damage identified. Therefore, the ceiling represented in the images were assigned severe damage. For Eq. (2), the ceiling was assigned component weight of 10 since the ceiling has the highest component weight. To assign the section significance as in Eq. (3), we detected the section represented in the image based on the objects detected inside it. For example, the first image has a refrigerator and an oven, then, the significance of the kitchen section is 3. The second image significance is 1 since we gave the highest significance to bathrooms and kitchens because of their water damage susceptibility.

### 6.3. Prototype website implementation

Next, we implemented the search and retrieval prototype website. We implemented a property database using the data model described in the previous section. We start by entering the importance weights of each component and the significance weights of each section. Then, we exported some of the currently available property information from different real estate websites and fed them to the Mask-RCNN damage detection network before we use them to fill in the database. We performed damage estimation and assigned each property a damage level. A Copy of the detected photos is saved in the property. We also developed a prototype real estate website[5] that pull the properties from the database showing the filter of the damage level as shown in Fig. 7.

## 7. Evaluation

Dunlap et al. (1988) were the first to conduct a study on measuring service quality in real estate. Grönroos (1984) explains that service quality is determined by two factors: technical quality and functional quality. Based on this, the evaluation procedure will be a two stage process. The Mask-RCNN Damage Level Detector network will be evaluated technically in the first stage. *Accuracy, training, and validation loss* will be the primary goals of this evaluation. The main objective of this evaluation is to assess the network's ability to identify damage and to examine how well it performs as a visual tool for real estate websites. An assessment of the prototype website's performance will be covered in the second stage. The *effectiveness and tangibility* of the suggested prototype will be demonstrated in this evaluation.

### 7.1. The Mask-RCNN evaluation

The accuracy, average precision, loss, and sensitivity of the Mask-RCNN will be the basis for the technical assessment. The damage level detection enhances image details or the concrete (tangible) aspects of service quality by providing essential semantic information about property images that involves the visual localization of damage. Table 1 shows the results from both the training and testing phases. The network will only identify a damage level if it is at least 0.85 confident in its accuracy. Examples of how the Mask-RCNN detects different Damage Level are shown in Fig. 8. Out of a total of 30 epochs, the highest validation accuracy was attained in the 24th epoch at 93.18 %. The network successfully identified ceiling damage with an average testing accuracy of 93.01 % and an average confidence level of 0.9201.

---

[4] MSCOCO dataset

[5] Prototype Website

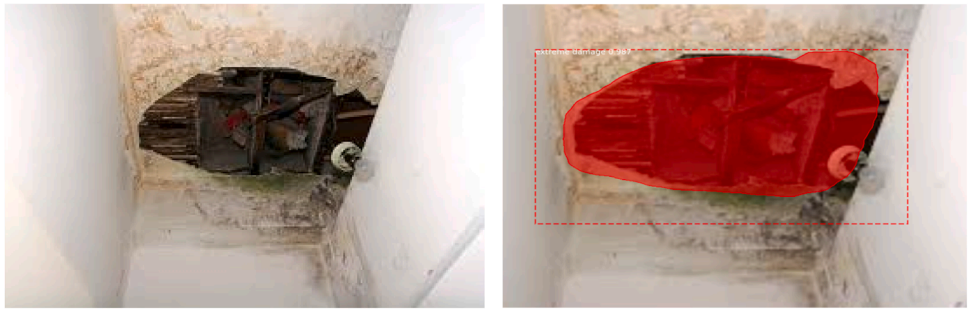

**Fig. 5.** Severe ceiling damage detection using Mask-RCNN showing instance segmentation.

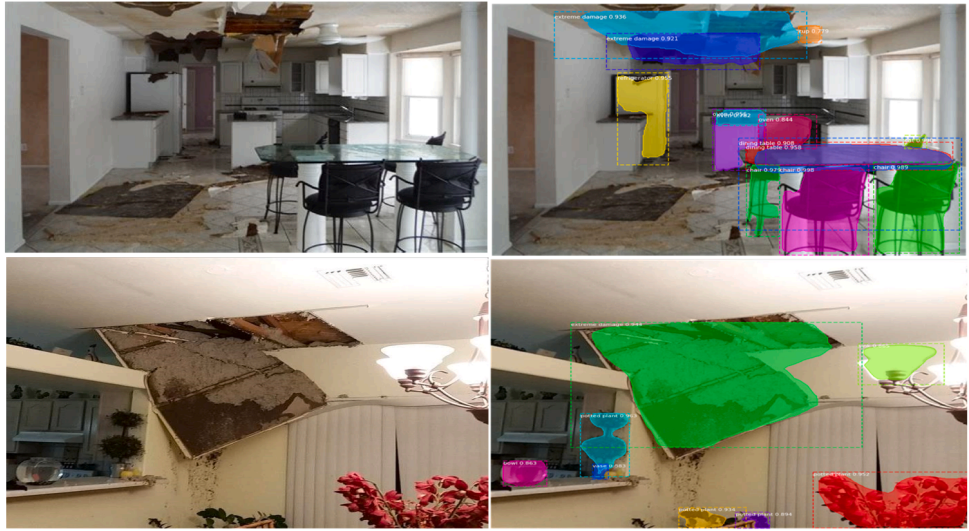

**Fig. 6.** Detecting damage including other objects in the image.

The analysis of the testing accuracy for each class was conducted by plotting it over a period of thirty epochs. As the sequence of epochs continued to increase, the average accuracy of the tests and the confidence of the Mapping-RCNN detection increased, as evidenced by Figs. 9 and 10. The highest level of test accuracy was seen at the 26th epoch, followed by a slight decrease at the following epochs. The steady increase in accuracy and the corresponding increase in confidence score suggest that the training and testing procedures of the network have been successful.

As some images fit into the severe and mild damage classes, respectively, it can be concluded that the latter is the most inaccurate and unstable class. Additionally, some images may be classified as either minor or mild damage at a lower level than the mild class.

### 7.1.1. The performance matrix of testing instances

To obtain a comprehensive assessment of the performance of each class, we constructed a confusion matrix to evaluate the precision and recall of the four classes, as discussed below.

1- **Recall**: Also referred to as *Sensitivity* or *Probability of Detection*, recall represents the fraction of correctly predicted positive samples out of all positive samples. Recall can be calculated using the following formula: TP/(TP+FN). The recall values for each class are as follows: Severe 0.871, Mild 0.946, Minor 0.948, and None class 0.941.
2- **Precision**: This metric indicates the proportion of positive class predictions that were correctly identified as positive. Precision can be calculated using the following formula: TP/(TP+FP). The precision values for each class are as follows: Severe 0.930, Mild 0.902, Minor 0.946, None 0.942.

### 7.1.2. Supplementary performance measures

Another important performance measure associated with deep learning networks is the loss training and validation loss. The loss is a representation of the cost function that needs to be reduced during training. The network evaluates the loss in the training and validation sets after every training epoch. The network's accuracy should be increasing with each passing epoch while the loss shoule be decreasing (Jancsary et al., 2012). Ensuring that the validation loss is a little bit lower than the training loss is a general rule of thumb. It indicates that the network is overfitting if the training loss is significantly greater than the validation loss. Fig. 11 shows a plot of the training and validation loss data over the course of the 30 training epochs. The training and validation loss begin at high values and decrease over time. There is also a small difference between the validation loss and the training loss in most of the 30 epochs, suggesting that the network has not experienced overfitting. The lowest validation loss occurred at the 26th epoch.

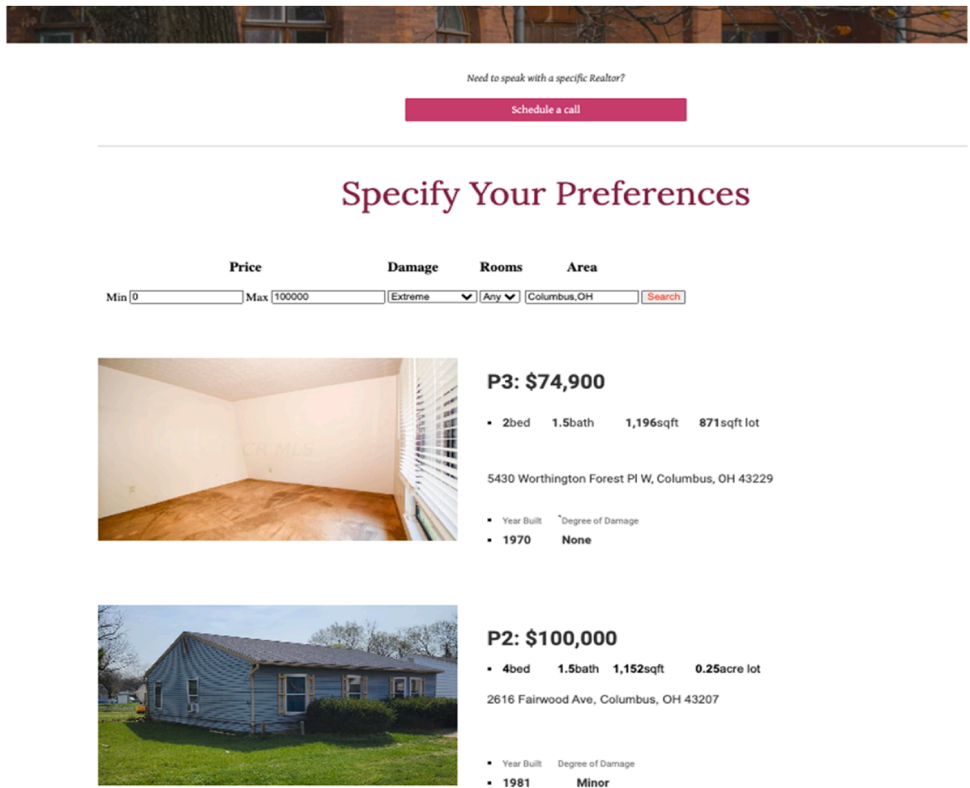

**Fig. 7.** Prototype real estate website that filter and display the properties with their damage level.

**Table 1**
Mask RCNN testing and validation accuracies at 85 % minimum confidence degree.

|              | Training instances | Validation instances | Test instances | Validation accuracy | Testing accuracy | Average confidence |
|--------------|--------------------|----------------------|----------------|---------------------|------------------|--------------------|
| Overall accuracy | 589 | 125 | 73 | 0.9314 | 0.9325 | 0.9321 |
| Severe class | 170 | 31 | 16 | 0.98324 | 0.9303 | 0.9734 |
| Mild class   | 201 | 53 | 21 | 0.9712 | 0.9089 | 0.8911 |
| Minor class  | 285 | 69 | 18 | 0.8278 | 0.9493 | 0.9325 |
| None class   | 131 | 29 | 19 | 0.9798 | 0.9436 | 0.9267 |

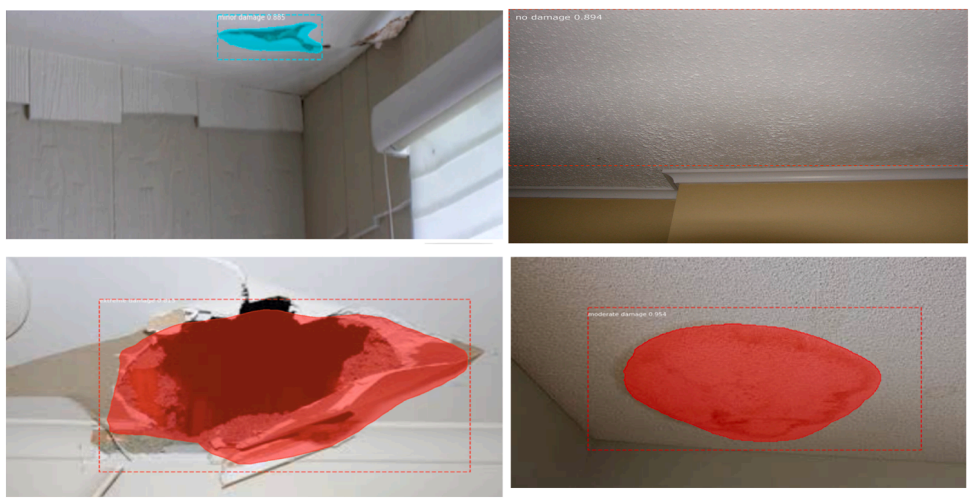

**Fig. 8.** Detection results for the for classes from the upper right: none, minor, mild, severe.

### 7.2. Evaluating the prototype website

The search and retrieval prototype will undergo a functional evaluation based on its *tangibility* and *efficiency*. Still in its prototype stage, we will not address the aesthetic or intangible aspects of service quality, such as linkage, appearance, and structure (Gronroos, 1988).

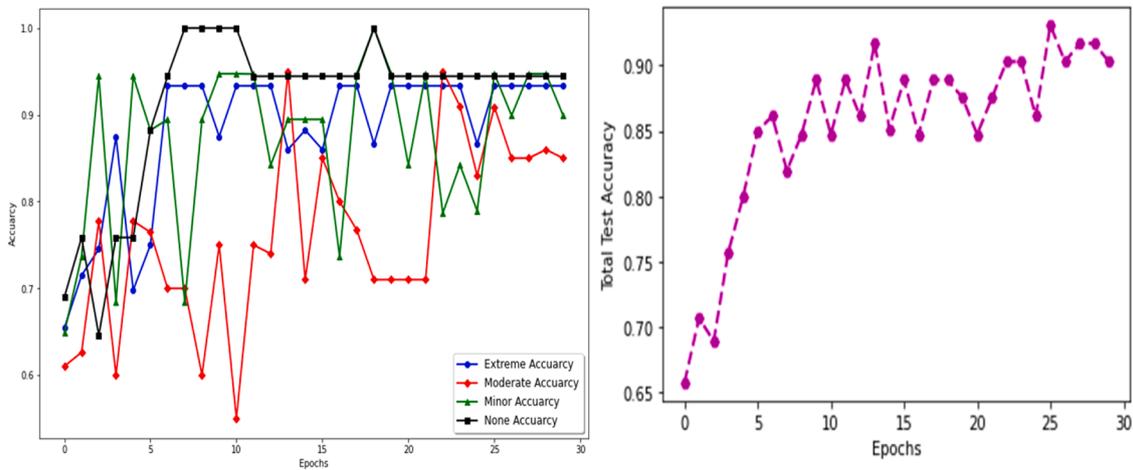

**Fig. 9.** Plotting the test accuracy of four classes vs. the overall test accuracy.

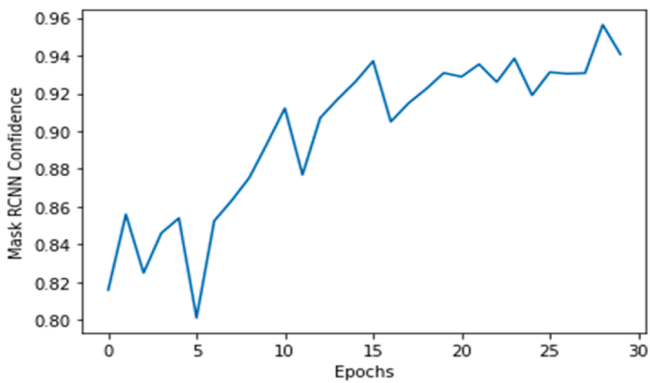

**Fig. 10.** Plotting the Mask-RCNN average confidence degree.

Instead, the focus is on the active service quality dimensions of tangibility, efficiency, and flexibility (Xu et al., 2013). The evaluation process will involve the use of illustrative scenarios to demonstrate the system's functionality and its ability to achieve its goals. Additionally, we will examine how the system improves the relevancy and efficiency of the search process.

### 7.2.1. Prototype website illustrative scenario

Real estate websites in the United States commonly utilize a centralized property database known as the MLS. The prototype website retrieves its listings from this MLS database, which can also be found on various well-known real estate websites. This website offers the functionality to filter properties based on the extent of damage. Additionally, it allows users to view detailed photos showcasing the damage, providing a more tangible experience with property images, as depicted in Fig. 12.

1- Imagine a house flipper who is looking for properties that are severely damaged. Accordingly, the flipper filtered the properties by selecting the severe damage level.
2- The website provided three properties, and when the ``show damage'' option was chosen, the images were displayed with the localizations of the damages.
3- The flipper chose property p4, which had the lowest price and the most rooms.
4- In order to assess *efficiency*, the flipper went to a well-known real estate website and entered the same parameters (price range, number of rooms, and location). When the extreme degree of damage was filtered, the listing for property p4 was the second search result to show up. On the other hand, listing P4, was the 17[th] listing on that well-known website. Instead of browsing through 17 listings, the flipper was able to find the desired property as the second listing

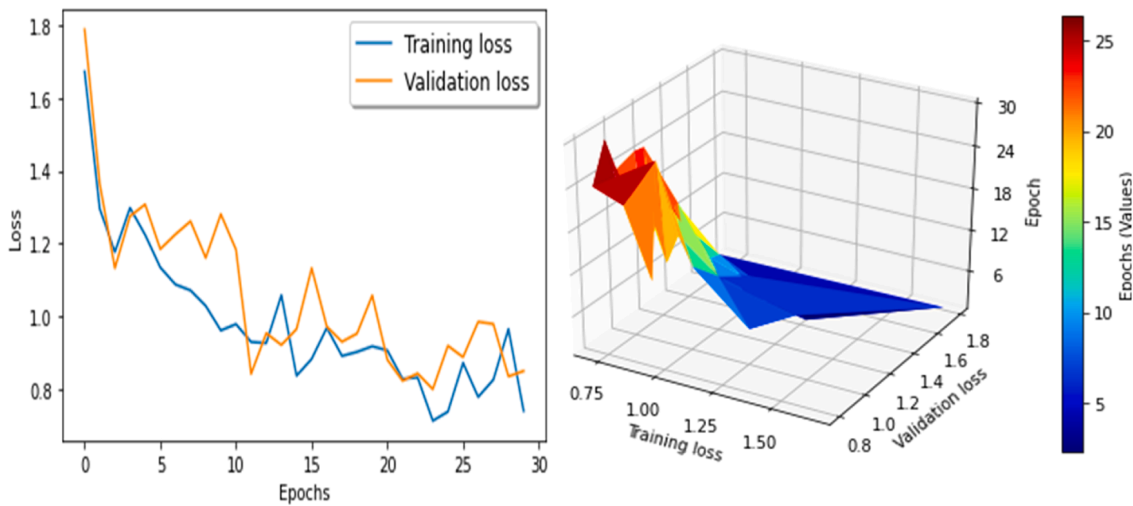

**Fig. 11.** Training loss vs. validation loss.

# Results

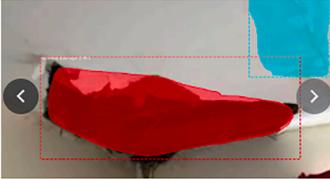

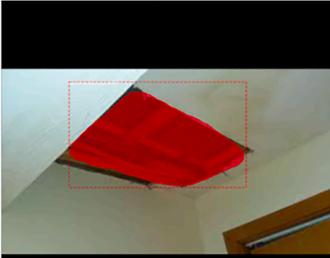

**Fig. 12.** Filtering results selecting severe damage level.

after using the *Damage Level* filter. This indicates that the addition of the *Damage Level* improved the effectiveness of the search procedure and made it easier for the flipper to locate the desired listing five times more quickly.

## 8. Discussion

Researchers have confirmed the significant effect of semantic and visual tools on enhancing e-service quality (Yuan et al., 2013; Ullah & Sepasgozar, 2019). For e-services that are primarily intangible, deep learning networks have the potential to improve the user experience. As a vital economic sector, this research introduced the *Damage Level* as a tangible feature to the real estate websites. By employing a trained Mask-RCNN, we can detect different levels of damage, which are assigned to each property's component, section, and then to the property overall. During network training, we concentrated on detecting ceiling damage. The ceiling is one of the most important components of any real estate section and is always given high importance because of its high maintenance cost. Since there is a lack of representative datasets for the *Damage Level*, we trained and tested images from web searches as well as real estate websites. In addition to the damage detection network, we also used an additional network to detect the section represented by objects and appliances in an image. For instance, an image with a toilet represents a bathroom. The trained network achieved 0.932 accuracy in detecting ceiling damage.

The evaluation of the trained Mask-RCNN network revealed that the severe damage and no damage classes achieved the highest levels of accuracy, while the mild damage class had the lowest accuracy. This is likely due to the fact that the mild damage class encompasses a wide range of damage levels, making it difficult to accurately classify. Some damages may fall into both the mild /minor or mild /severe categories. Upon closer examination, it was found that dark moldy spots are often identified as severe damage rather than mild. Therefore, the expertise of real estate professionals is necessary to define the damage spectrum for each *Damage Level*. In contrast, the none and severe damage classes achieved the highest levels of accuracy because they represent the two ends of the damage spectrum. To demonstrate the functionality of the system, we developed a prototype website that allows users to filter properties based on their *Damage Level* and view detailed information about the damages. This visual feature of the *Damage Level* not only enhances the *tangibility* of real estate photos but also facilitates a more *efficient* search process, enabling users to find their desired listings more quickly.

## 9. Limitations and future work

The Mask-RCNN damage detection network achieved an accuracy of 93 % in identifying damages. However, determining whether the damage is minor, mild, or severe poses significant challenges and is subjective. Some damages may even fall into multiple damage categories. For instance, a large moldy spot could be classified as both mild and severe damage. To overcome these challenges, the involvement of real estate specialists is necessary to establish specific criteria for each damage category.

The number of training, testing, and validation samples needed to determine the *Damage Level* in the ceiling was constrained due to the difficulty in obtaining a pertinent dataset for real estate damages. For a thorough examination of a property, more training samples are needed to identify damage in various components and sections.

Ceiling elements like air vents, smoke detectors, speakers, and light fixtures should be accounted for as supplementary classes during the training phase. To achieve thorough and accurate damage detection, it is essential to incorporate all fixtures present in real estate images into the training process. This undertaking necessitates a substantial quantity of training samples for each individual object.

Future work will entail the creation of a comprehensive dataset that captures various levels of damage to real estate. The dataset must encompass all components and sections of real estate in different states of disrepair. It is also important to consider the source of the damage, such as water leaks or paint cracks, when quantifying the extent of the damage. Even for properties with the same damage level, the source of the damage can significantly impact the cost of repairs. For instance, a mild damage resulting from paint crack may be less expensive to rectify than mild damage caused by areas affected by mold. However, this phase necessitates the expertise of real estate inspectors and specialists who can accurately identify the sources of damage.

The impact of the proposed model on practice includes enhancing the service quality of real estate websites through enhancing *tangibility and efficiency*. Moreover, the introduction of new visual descriptive features such as the *Damage Level* is enhancing the *information quality* provided to users to make informed decisions. The work could be extended to other e-services in dire need of more tangible descriptive features for enhanced service quality.

This research makes several contributions to theory. It emphasizes the importance of incorporating semantic visual features such as the *Damage Level*. The research also highlights the impact of using the new technologies of deep learning to enhance the tangibility and efficiency of e-services. The "*tangible*'' dimensions of service quality have long been ignored, despite their significant impact on enhancing the economic outcomes of e-services. The research suggests examining the impact of other visual descriptive features such as material look and feel (Yuan et al., 2017) on other vital financial sectors such as retail, education, and cosmetics.

## 10. Conclusion

E-services usually lack the semantic tangible features of products and services, affecting their service quality. Deep learning for computer vision could be used to detect semantic features that enhance the tangibility of e-services. Moreover, these features could be used as search filters to improve the search process efficiency of e-services websites. In application, we introduced the *Damage Level* feature, which estimates a property's state of maintenance. We trained a Mask-RCNN network to detect and allocate damages in real estate images. In addition, we assigned each property a *Damage Level* based on the damages of its components and sections. The *Damage Level* is used later in a prototype website as a search filter to provide a visual description of the property's state of maintenance. Introducing the Damage Level will not only enhance the *tangibility* of real estate websites but also enhance *efficiency* by filtering properties based on the specified damage level. The research application could be extended to other e-services by using deep learning to detect other descriptive features in many vital e-services, such as retail and cosmetics.

**CRediT authorship contribution statement**

**Samaa Elnagar:** Conceptualization, Data curation, Formal analysis, Methodology, Resources, Software, Visualization, Writing – original draft, Writing – review & editing.

**Declaration of competing interest**

The authors declare that they have no known competing financial interests or personal relationships that could have appeared to influence the work reported in this paper.

**Data availability**

Data will be made available on request.

## Appendix

*Property database*

To gain a deeper understanding of how the data flow in the proposed system and how the Degree of Damage will be reflected in the real estate data records, the system data model is expressed in Fig. A1 and explained below in detail.

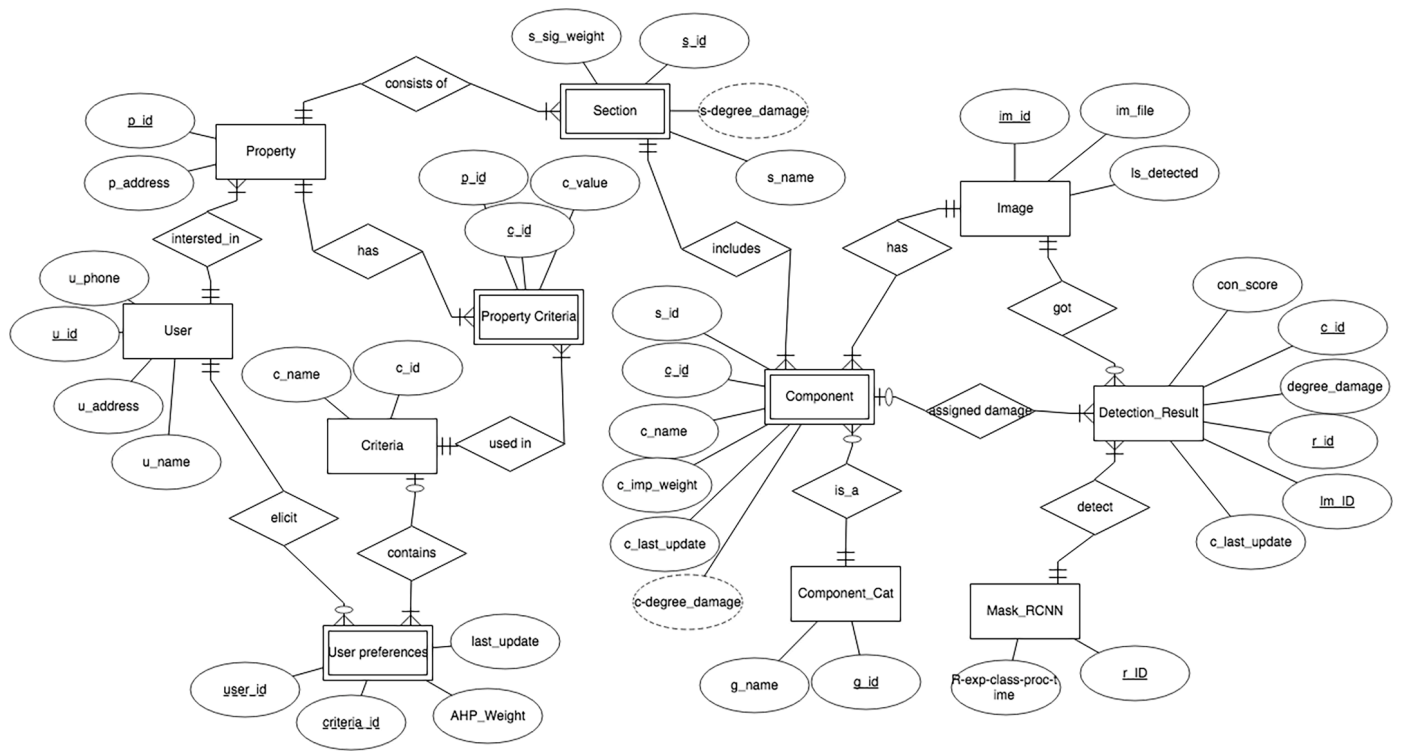

**Fig. A1.** The data model for the real estate search system.

1. **Property:** each property has an id, address and a set of different criteria pulled from the **Criteria** entity.
2. **Criteria:** it is a lookup table that contains a set of property's criteria such as *no_floors, price,* and *sqft.*
3. **Property Criteria:** contains the values for each of the property's criteria.
4. **User:** this entity contains the information of the users, such as email, phone, etc.
5. **User Preferences:** stores the user's preference weights of each criterion in comparison to other criteria. The *AHP_Weight* is the weight generated from the AHP for each criterion.
6. **Section:** or property's sections such as bedrooms, living rooms; the section is dependent on property.
   - *s_sig_weight*: The significance score of the section based on its susceptibility to damage. The values of *s_ degree_damage* is derived from the sum of the damage of its components.
7. **Component:** is each part of a section such as the *Ceiling, Floor, Door, Closet* of a section; component is dependent on section (relative to the section).
   - *c_imp_weight*: real estate experts assign a relative weight to each component, usually based on its cost of maintenance.
   - *c_degree_damge*: is the value assigned by the last damage estimation, and it would be Null if the component isn't previously estimated.
   - *c_Last_Update*: is the date of the last damage estimation or Null if it is not damage estimated.
8. **Comp_Category:** is the components lookup (e.g., Ceiling, Floor, Door, Closet, etc.). This entity is independent of any particular component.
9. **Mask-RCNN_Model:** stores the configurations of the Mask-RCNN models running on different computing units.
10. *R_Exp_Class_Proc_Time*: is the average time spent in processing each of the four damage classes.
11. **Image:** each image contains one or more components of a section. Each image has a name and file location. *Is_detected* marks undetected images to be detected when the Mask-RCNN_Models are idle.
12. **Detection_Result:** for each detected component in the image, the *degree_damge* and *con_score* are saved. *con_score* is the confidence score that Mask-RCNN provides for each detected component.

### Further Reading